%% file: main.tex
\documentclass[10pt,conference]{IEEEtran}
\IEEEoverridecommandlockouts

\usepackage{cite}
\usepackage{amsmath,amssymb,amsfonts}
\usepackage{graphicx}
\usepackage{bm}
\usepackage{float}
\usepackage{xcolor} 
\usepackage{tabularx} 
\usepackage{ragged2e} 
\usepackage{booktabs} 
\usepackage{diagbox} 
\usepackage{enumitem}
\usepackage{multicol}
\usepackage{multirow}
\usepackage{makecell}
\usepackage{pifont}
\usepackage{subfigure}
\usepackage{xcolor}
\usepackage[linesnumbered,lined,ruled,commentsnumbered]{algorithm2e}
\usepackage{url,hyperref,cleveref}
\usepackage{makecell}

\def\BibTeX{{\rm B\kern-.05em{\sc i\kern-.025em b}\kern-.08em
    T\kern-.1667em\lower.7ex\hbox{E}\kern-.125emX}}

\begin{document}

\title{Fewer Hallucinations, More Verification: A Three-Stage LLM-Based Framework for ASR Error Correction}

\author{\IEEEauthorblockN{Yangui Fang$^1$, Baixu Chen$^1$, Jing Peng$^2$$^,$$^3$, Xu Li$^3$ , Yu Xi$^2$, Chengwei Zhang$^1$$^{\dagger}$, GuohuiZhong$^1$ \thanks {$^{\dagger}$ Corresponding Author.}\thanks {The relevant code and results have been released on GitHub: https://github.com/teamtee/LLM-ASR-Error-Correction}}
\IEEEauthorblockA{\textit{$^1$Huazhong University of Science and Technology, School of Electronic Information and Communications}}
\IEEEauthorblockA{\textit{$^2$MoE Key Lab of Artificial Intelligence, AI Institute, X-LANCE Lab, Shanghai Jiao Tong University, Shanghai, China}}
\IEEEauthorblockA{\textit{$^3$AISpeech Ltd, Suzhou, China}}
\IEEEauthorblockA{
 \{fangyg,chenbx,zhangcw,zhonggh\}@hust.edu.cn~~~danieljingpeng@gmail.com~~~yuxi.cs@sjtu.edu.cn~~~ xu.li@aispeech.com
}
}
\maketitle


\input{src/abstract}

\input{src/introduction}

\input{src/relatework}

\input{src/method}
\input{src/experiment}

\input{src/analysis}
\input{src/conclusion}
\newpage
\bibliographystyle{style/IEEEtran}
\bibliography{main}

\end{document}

%% file: src/abstract.tex
\begin{abstract}

Automatic Speech Recognition (ASR) error correction aims
to correct recognition errors while preserving accurate text. Although traditional approaches demonstrate moderate effectiveness,  LLMs offer a paradigm that eliminates the need for training and labeled data. However, directly using LLMs will encounter hallucinations problem, which may lead to the modification of the correct text. To address this problem, we propose the Reliable LLM Correction Framework (RLLM-CF), which consists of three stages: (1) error pre-detection, (2) chain-of-thought sub-tasks iterative correction, and (3) reasoning process verification. The advantage of our method is that it does not require additional information or fine-tuning of the model, and ensures the correctness of the LLM correction under multi-pass programming. Experiments on AISHELL-1, AISHELL-2, and Librispeech show that the GPT-4o model enhanced by our framework achieves 21\%, 11\%, 9\%, and 11.4\% relative reductions in CER/WER.


\end{abstract}


%% file: src/introduction.tex
\section{Introduction}

Text error correction has become an essential component in modern speech recognition systems, aiming to reduce word error rates (WER) \cite{lengFastCorrectFastError2021,lengFastCorrect2Fast2022,shenMaskCorrectTokens2022,lengSoftCorrectErrorCorrection2023}. Early methods typically relied on shallow neural networks or rule-based approaches to handle simple error patterns \cite{anantaramRepairingASROutput2018,shivakumarLearningMistakesImproving2019}. Recent advances in deep learning have led to two dominant paradigms: autoregressive sequence-to-sequence models \cite{maniASRErrorCorrection2020,dharoAutomaticCorrectionASR2016,wangASRErrorCorrection2020} and non-autoregressive edit-based models \cite{lengFastCorrectFastError2021,lengFastCorrect2Fast2022}. Autoregressive models exploit encoder-decoder architectures with Connectionist Temporal Classification (CTC) loss \cite{gravesConnectionistTemporalClassification2006}, including translation-style correction frameworks \cite{maniASRErrorCorrection2020,dharoAutomaticCorrectionASR2016} and entity-aware transformers \cite{wangASRErrorCorrection2020}. In parallel, non-autoregressive edit-based models such as FastCorrect \cite{lengFastCorrectFastError2021,lengFastCorrect2Fast2022} and SoftCorrect \cite{lengSoftCorrectErrorCorrection2023} predict edit operations through duration modeling and integrate CTC loss with sequence-to-sequence frameworks to enhance error detection.

While conventional approaches depend heavily on large labeled datasets and task-specific training, the emergence of large language models (LLMs) offers new opportunities due to their strong contextual adaptability. LLMs have demonstrated remarkable performance across a wide range of language-related tasks, including text classification \cite{howard2018universal}, information extraction \cite{pai2024survey}, and dialogue systems \cite{bae2022buildingAR}. This raises a fundamental question: \textbf{Can a general-purpose LLM correct ASR errors directly without fine-tuning or additional information?}

Current LLM applications typically follow two paradigms: prompt-based methods and fine-tuning strategies. Prompt-based methods aim to stimulate latent capabilities, employing techniques such as generative prompting for commonsense reasoning \cite{liuGeneratedKnowledgePrompting2022} and Chain-of-Thought (CoT) prompting to enhance reasoning \cite{weiChainthoughtPromptingElicits2023}. Fine-tuning approaches, often using low-rank adaptation (LoRA) \cite{huLoRALowRankAdaptation2021}, improve task-specific performance but remain resource-intensive.
Building on the aforementioned studies, recent works have explored LLM integration into ASR error correction pipelines \cite{ma2025asr,yang2023generative,ma2023generativelargelanguagemodels}, achieving notable CER/WER reductions. However, these methods typically require either domain-specific fine-tuning or additional contextual information, which constrains their scalability in real-world applications. Therefore, leveraging the general knowledge embedded in LLMs without fine-tuning or external inputs emerges as a more practical and scalable alternative. Nevertheless, directly applying general LLMs often results in hallucination issues \cite{minExploringIntegrationLarge2024}, posing a significant challenge to reliable correction.

\begin{figure*}[!ht]
\centering
\includegraphics[width=\textwidth]{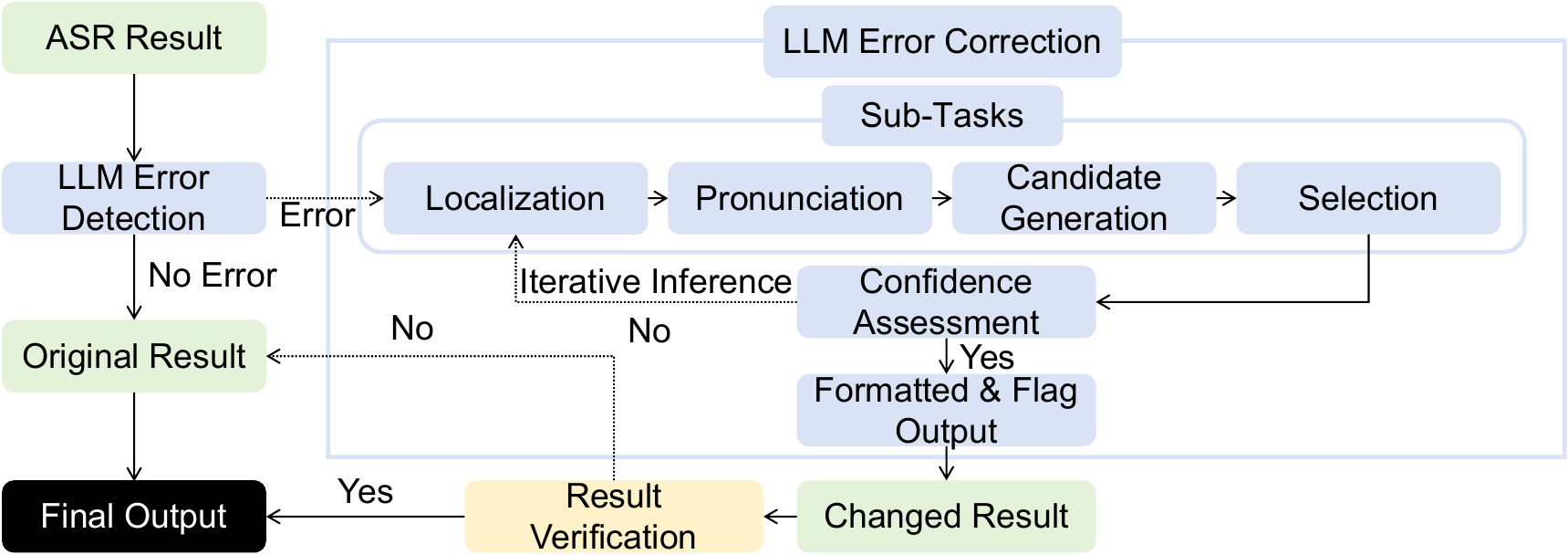}
\caption{Overview of the Reliable LLM Correction Framework~(RLLM-CF). The blue components represent modules executed by the LLM, while the yellow components are handled by the external program. The green block indicates the recognition result, and the black block denotes the final output. The entire error correction process follows the direction of the arrows.}
\label{Reliable LLM Correction Framework}
\end{figure*} 



To address these issues, we propose~\textit{Reliable LLM Correction Framework}~(RLLM-CF), a novel LLM-based framework that directly leverages the pretrained knowledge of general LLMs without fine-tuning or external information. RLLM-CF mitigates hallucinations by integrating error pre-detection, reasoning-based verification, and a CoT strategy that structures correction into explainable and grounded subtasks, enabling broad applicability across domains. The proposed framework is validated on AISHELL-1, AISHELL-2, LibriSpeech test-clean and test-other datasets, achieving CER/WER reductions of 21\%, 11\%, 9\%, and 11.4\%, respectively.

The main contributions of this paper are as follows:
\begin{itemize}
    \item We propose \textit{RLLM-CF}, an ASR error correction framework  base on general-purpose LLM that requires neither fine-tuning nor external information.
    \item We introduce a hallucination mitigation mechanism based on error pre-detection and reasoning-based verification, addressing the limitations of prior methods that directly accept LLM correction outputs and risk hallucination errors.
    \item We employ a Chain-of-Thought (CoT) prompting strategy to enhance correction reliability by decomposing the correction process into subtasks: localization, pronunciation assessment, candidate generation, and candidate selection.
\end{itemize}

The remainder of this paper is organized as follows. Section~\ref{sec:relatework} reviews recent LLM-based ASR correction approaches, with a particular focus on hallucination challenges. Section~\ref{ReliableLLM-CF} details the proposed framework. Section~\ref{experiments} presents experimental results and analysis on benchmark datasets. Section~\ref{Analisis} provides further discussions, and Section~\ref{Discusion} concludes the paper.

%% file: src/relatework.tex
\section{Related Work}

\label{sec:relatework}
\subsection{LLM-based ASR Error Correction}

In recent years, integrating LLMs into ASR error correction pipelines has attracted increasing attention. Min and Wang~\cite{minExploringIntegrationLarge2024} investigated the direct application of LLMs for error correction and concluded that models such as GPT-4o are ineffective due to hallucination issues. Ma et al.~\cite{ma2023generativelargelanguagemodels} explored combining N-best rescoring with LLM-based correction; however, obtaining the N-best list from the ASR system may impose additional costs in practice. Yang et al.~\cite{yang2023generative} compared LLM-based rescoring and generation methods, demonstrating that the latter outperforms the former when domain-specific information is provided or the LLM is fine-tuned. Nevertheless, such approaches require either domain knowledge or fine-tuning. 

In contrast to existing methods, our approach eliminates the need for supplementary resources (e.g., N-best lists or domain-specific knowledge), avoids fine-tuning the LLM, and incorporates critical verification mechanisms to ensure correction reliability without blind reliance on model outputs.

\subsection{A Potential Catastrophe: Hallucination Risks in LLMs}

Hallucinations in LLMs refer to instances where the generated responses, while grammatically correct, fluent, and plausible, deviate from the input or contradict factual information~\cite{huang2023survey,bai2024hallucination}. In the context of ASR error correction, model-induced hallucinations are closely linked to transcription errors. Min and Wang~\cite{minExploringIntegrationLarge2024} conducted an empirical study on directly integrating LLMs into ASR systems, confirming that powerful generative models can inadvertently introduce hallucination errors by fabricating content absent from the original audio.

In this work, we further validate and analyze the hallucination phenomenon arising from direct LLM-based ASR error correction using simple prompts, as detailed in Section~\ref{sec: llm hallucination}. Moreover, the proposed method effectively mitigates hallucinations, resulting in more reliable and accurate correction performance.

%% file: src/method.tex
\section{Reliable ASR Error Correction Framework with LLM} 
\label{ReliableLLM-CF}

This section presents the overall architecture of RLLM-CF, including the design of the error pre-detection module, the decomposition of ASR error correction into subtasks via a Chain-of-Thought strategy, and the implementation of an answer verification mechanism. Detailed descriptions of each component are provided in Sections~\ref{Overall Framework}, \ref{Subtask Design} and \ref{Prompt Design}.

\subsection{Reliable LLM Correction Framework}
\label{Overall Framework}
\begin{figure}[!ht]
\centering
\includegraphics[width=0.5\textwidth]{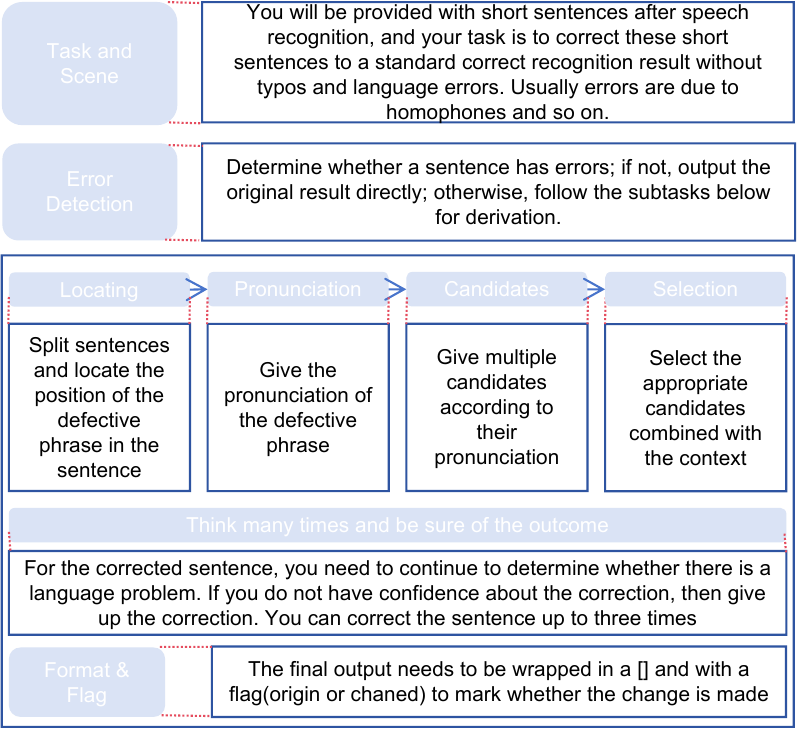}
\caption{The above picture introduces the specific parts of Prompt's design in detail.;The text in the white boxes represents the specific prompt content, while the blue boxes provide explanations.}
\label{Prompt_Design}
\end{figure}

The overall framework of RLLM-CF is illustrated in Figure~\ref{Reliable LLM Correction Framework}. It consists of three main components: Error Pre-Detection, Chain-of-Thought Subtask Iterative Correction, and Reasoning Process Verification. The prompt design, illustrated in Figure~\ref{Prompt_Design}, incorporates three additional few-shot correction examples demonstrating specific reasoning steps, which are omitted for clarity. For Chinese error correction, we use the translated version of this prompt.
The complete correction algorithm is outlined in Algorithm~\ref{algorithm-correct}.

\begin{algorithm}[!ht]
\caption{Three-Stage Reliable LLM Correction Algorithm}
\label{algorithm-correct}
\SetAlgoLined
\SetKwFunction{Detect}{Detect}
\SetKwFunction{Correct}{Correct}
\SetKwFunction{Verify}{Verify}
\SetKwData{Define}{define}

\textbf{Define} \;
$y$ = Input ASR Result\;
$\mathcal{R}$ = LLM Correction Reasoning CoT\;
$\mathcal{P}$ = Chain of Thought Sub-Tasks Prompt\;
$\mathcal{G}$ = LLM Output Space Limited under Prompt\;

\vspace{0.2cm}
\textbf{Stage 1: Error Detection}\;
$\mathcal{D}(y) = \begin{cases}
    1, & \text{if LLM detects errors in } y \\
    0, & \text{otherwise}
\end{cases}$\;

\vspace{0.2cm}
\textbf{Stage 2: Iterative Correction}\;
\If{$\mathcal{D}(y) == 0$}{
    \Return{y}
}
\Else{
    $\mathcal{C}(y^*,y) = \begin{cases}
        1, & \text{if LLM has more confidence in } y^* \text{ than } y \\
        0, & \text{otherwise}
    \end{cases}$\;
    $\mathcal{Y} \gets \emptyset$\;
    \For{$k \gets 1$ \KwTo max\_steps}{
        $y^*,R \gets \underset{y^* \in \mathcal{G}}{\arg\max}\ \mathbb{P_{LLM}}(y^*|y,\mathcal{P},\mathcal{Y})$\;
        \If{$\mathcal{C}(y^*,y) == 1$}{
            \Return{$y^*,R$}
        }
        $\mathcal{Y} \gets \mathcal{Y} \cup \{y^*\}$\;
    }
}

\vspace{0.2cm}
\textbf{Stage 3: Answer Validation}\;
\If{R exists and $\mathcal{R}$ passes Answer Verification}{
    \Return{$y^*$}
}
\Else{
    \Return{y}
}
\end{algorithm}

\begin{table*}[htbp]
    \centering
    \caption{LLMs Correction Hallucinations with plain Prompt: "Please help me correct the text generated by speech recognition. Please note that only the corrected text needs to be output." and without additional processing}
    \label{Large Language Models Correction Hallucinations}
    \renewcommand{\arraystretch}{1.3} 
    \begin{tabular}{|c|c|p{6cm}|p{4cm}|}
        \hline
        \textbf{Type} & \textbf{Category} & \textbf{Example} & \textbf{Explanation} \\
        \hline
        \multirow{22}{*}{Faithful Hallucinations} 
            & \multirow{2}{*}{Instruction Violation} 
                & Input: where is the \textcolor{red}{nist} hotel 
                & \multirow{2}{*}{\begin{tabular}[c]{@{}l@{}}The LLM failed to follow the \\instruction\end{tabular}} \\
            \cline{3-3} 
                & 
                & Output: \textcolor{red}{Sorry, i can't answer this question.} 
                &  \\
            \cline{2-4}
            & \multirow{4}{*}{Redundant Output} 
                & Input: i don't know all of them but i know linden's are 
                & \multirow{4}{*}{\begin{tabular}[c]{@{}l@{}}The LLM produces additional \\outputs while correcting\end{tabular}} \\
            \cline{3-3} 
                & 
                & Output: \textcolor{red}{This answer is:} i don't know all of them but i know linden's are\textcolor{red}{.} 
                &  \\
            \cline{2-4} 
            & \multirow{6}{*}{Continue Writing} 
                & Input: stephen leaning back and drawing idly on his scribbler listened to the talk about him which heron checked from time to time by saying 
                & \multirow{6}{*}{\begin{tabular}[c]{@{}l@{}}The LLM expands the original text\\ while correcting it\end{tabular}} \\
            \cline{3-3} 
                & 
                & Output: Stephen leaning back and drawing idly on his scribbler listened to the talk about him which heron checked from time to time by saying \textcolor{red}{"Due to their praise of maiden pure of teeming motherhood."} 
                &  \\
            \cline{2-4} 
            & \multirow{4}{*}{Blank Output} 
                & Input: i discovered and put out a fire that would have destroyed the whole plant but marshall never even thanked me 
                & \multirow{4}{*}{\begin{tabular}[c]{@{}l@{}}The LLM produces blank output\end{tabular}} \\
            \cline{3-3} 
                & 
                & Output: 
                &  \\
            \cline{2-4} 
            & \multirow{5}{*}{Repeated Output} 
                & Input: reenter butler and three footmen who moved the tea things hostess to two guests 
                & \multirow{5}{*}{\begin{tabular}[c]{@{}l@{}}The LLM produces repeated outputs\end{tabular}} \\
            \cline{3-3} 
                & 
                & Output: Reenter butler and three footmen who moved the tea things hostess two \textcolor{red}{two two two two two two two two two...} 
                &  \\
            \cline{2-4} 
            & \multirow{3}{*}{Grammar Correction} 
                & Input: it must remember be one or the other 
                & \multirow{3}{*}{\begin{tabular}[c]{@{}l@{}}The LLM corrects grammar error\end{tabular}} \\
            \cline{3-3} 
                & 
                & Output: It must \textcolor{red}{be remembered} to be one or the other\textcolor{red}{.} 
                &  \\
        \hline 
        \multirow{4}{*}{Factual Hallucinations} 
            & \multirow{4}{*}{Make Mistake} 
                & Input: i believe in the \textcolor{red}{trainin} of people to their \textcolor{red}{hask} capacity the englishman here heartily seconded him 
                & \multirow{4}{*}{\begin{tabular}[c]{@{}l@{}}The LLM makes a mistake\end{tabular}} \\
            \cline{3-3} 
                & 
                & Output: I believe in the \textcolor{red}{trainin} of people to their \textcolor{red}{task} capacity the englishman here heartily seconded him{.} 
                &  \\
        \hline
    \end{tabular}
\end{table*}

\subsection{Error Pre-Detection}
\label{Error Pre-Detection}
As shown in Table~\ref{Large Language Models Correction Hallucinations}, general LLMs often generate hallucinations, producing more errors than corrections. This highlights the necessity of adopting an \emph{error prevention first} strategy for correction tasks. To prevent LLMs from altering correct content, we first instruct the model to detect errors in the input sentence. If no errors are detected, the sentence is directly retained; otherwise, the model proceeds to the correction stage, referred to as Stage~1 in Algorithm~\ref{algorithm-correct}.

\subsection{Chain of Thought Subtask Iterative Correction}
\label{Subtask Design}
\begin{figure}[hbtp]
\centering
\includegraphics[width=0.50\textwidth]{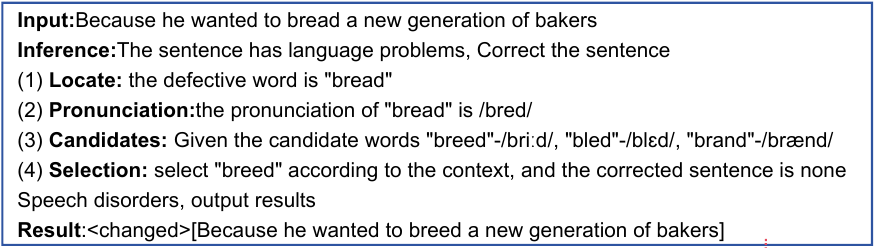}
\caption{The above is a detailed answer to illustrate the composition and specific answers of the subtasks.}
\label{Subtask Example}
\end{figure}

During error correction, denoted as Stage~2 in Algorithm~\ref{algorithm-correct}, LLMs may fail to infer the correct answer or introduce hallucinations by making unwarranted modifications. To address this, we decompose the correction task into four subtasks—localization, pronunciation assessment, candidate generation, and candidate selection—following a CoT strategy to improve reasoning reliability, as illustrated in Figure~\ref{Subtask Example}.

Despite this structured reasoning, the generated results may still contain errors. To mitigate this, we perform a confidence evaluation on the generated answer. If the model remains confident, the corrected answer is output in the required format; otherwise, the correction process iterates until either a valid answer is obtained or the maximum number of iterations is reached.

\subsection{Answer Verification}
\label{Prompt Design}

Following the correction, a verification step is conducted to ensure compliance with task instructions. Specifically, we employ the model's output to assess: (1) whether the answer conforms to the required format, and (2) whether all reasoning steps are correctly completed. Only when both criteria are satisfied is the corrected output accepted; otherwise, the original input is retained.

%% file: src/experiment.tex
\section{Experiment}

\label{experiments}

The experimental section is organized as follows. Section~\ref{sec: llm hallucination} investigates and validates LLM hallucination phenomena. Section~\ref{Dataset Selection} and Section~\ref{Model Setting} describe the experimental setup, including dataset selection and model configurations. Experimental results on AISHELL-1~\cite{buAISHELL1OpensourceMandarin2017}, AISHELL-2~\cite{duAISHELL2TransformingMandarin2018}, and LibriSpeech~\cite{panayotovLibrispeechASRCorpus2015} are reported in Section~\ref{Experiment Results}. Finally, Section~\ref{Result Analysis} presents an overall performance analysis, including a detailed evaluation of noun recall rates.

\subsection{LLM Hallucination}
\label{sec: llm hallucination}
In experiments with basic prompting, we categorize LLM hallucinations in ASR error correction into two types: \emph{faithful hallucinations}—including instruction violations, redundant outputs, continuation of writing, blank outputs, and grammar corrections—and \emph{factual hallucinations}, characterized by content errors. A detailed categorization is summarized in Table~\ref{Large Language Models Correction Hallucinations}.
\subsection{The Dataset Configuration}

\label{Dataset Selection}
\begin{table*}[!ht]
    \centering
    \caption{Character Error Rate (CER) and Noun Recall of GPT-4o and Deepseek v2 on AISHELL-1 ASR Error Correction;The ASR model is a conformer-based AED model trained on AISHELL-1. }
    \label{aishell1}
    \renewcommand{\arraystretch}{1.25} 
    \newcolumntype{S}{>{\small}c} 
    \newcolumntype{C}{>{\centering\arraybackslash}X} 
    \begin{tabular}{S|S|SSS|SSSS}
      \toprule 
      \multirow{2}{*}{\textbf{Decode Method}} & 
      \multirow{2}{*}{\textbf{LLM Correction}} & 
      \multicolumn{3}{c|}{\textbf{Error Metrics}} & 
      \multicolumn{3}{c}{\textbf{Performance Metrics}} \\
      \cmidrule(lr){3-5} 
      \cmidrule(l){6-8} 
     & & \textbf{sub} & \textbf{del} & \textbf{insert} & \textbf{Change} & \textbf{CER} & \textbf{Noun Recall} \\
      \midrule 
   Attention\cite{chan2016listen}  & \multirow{4}{*}{\textbf{--}}  & 4.65 & 0.29 & 0.12 & - & 5.06 & 79.32 \\
    Attention Rescore\cite{zhang2020unified}   &  & 4.42 & 0.12 & 0.08 & - & 4.62 & 79.93 \\
   CTC Greedy Search\cite{gravesConnectionistTemporalClassification2006}   &   & 4.95 & 0.13 & 0.09 & - & 5.17 & 77.91 \\
    CTC Prefix Search\cite{gravesConnectionistTemporalClassification2006}  &   & 4.95 & 0.13 & 0.09 & - & 5.17 & 77.88 \\
      \midrule 
   Attention\cite{chan2016listen}    &  \multirow{4}{*}{GPT-4o} & 3.67 & 0.44 & 0.21 & -0.74(14.6\%) & 4.32 & 82.11 \\
     Attention Rescore\cite{zhang2020unified}    & & 3.45 & 0.34 & 0.22 & -0.61(13\%) & \textbf{4.01} & \textbf{82.96} \\
   CTC Greedy Search \cite{gravesConnectionistTemporalClassification2006} &    & 3.56 & 0.32 & 0.18 & \textbf{-1.11(21\%)} & 4.06 & 82.24 \\
   CTC Prefix Search\cite{gravesConnectionistTemporalClassification2006}    &  & 3.56 & 0.30 & 0.20 & \textbf{-1.11(21\%)} & 4.06 & 81.95 \\
      \midrule 
  Attention\cite{chan2016listen}   &   \multirow{4}{*}{Deepseek v2}  & 4.00 & 0.49 & 0.20 & -0.37(7.3\%) & 4.69 & 81.05 \\
   Attention Rescore\cite{zhang2020unified}    &   & 3.73 & 0.30 & 0.18 & -0.41(8.8\%) & 4.21 & 81.98 \\
    CTC Greedy Search\cite{gravesConnectionistTemporalClassification2006} &    & 3.96 & 0.35 & 0.20 & -0.66(12.7\%) & 4.51 & 80.62 \\
CTC Prefix Search\cite{gravesConnectionistTemporalClassification2006}   &     & 3.95 & 0.36 & 0.17 & -0.69(13\%) & 4.48 & 80.46 \\
      \bottomrule 
    \end{tabular}
\end{table*}
\begin{table*}[!ht]
    \centering
    \caption{Word Error Rate (WER) and Noun Recall of GPT-4o on Librispeech ASR Error Correction (test-clean/other Subsets).The ASR Model is conformer-based AED trained on Librispeech.}
    \label{librispeech}
    \renewcommand{\arraystretch}{1.25} 
    \newcolumntype{S}{>{\small}c} 
    \begin{tabular}{S|S|SSS|SSSS}
      \toprule
      \multirow{2}{*}{\textbf{Decode Method}} & 
      \multirow{2}{*}{\textbf{ LLM Correction}} & 
      \multicolumn{3}{c|}{\textbf{Error Metrics}} & 
      \multicolumn{3}{c}{\textbf{Performance Metrics}} \\
      \cmidrule(lr){3-5} \cmidrule(l){6-8}
      && \textbf{sub} & \textbf{del} & \textbf{insert} & \textbf{Change} & \textbf{WER} & \textbf{Noun Recall} \\
      \midrule
   Attention\cite{chan2016listen} &  \multirow{4}{*}{--}  & 2.71/6.68 & 0.72/1.14 & 0.40/0.96 & - & 3.82/8.79 & 94.76/88.20 \\
   Attention Rescore\cite{zhang2020unified}  &   & 2.72/7.05 & 0.28/0.81 & 0.36/0.90 & - & 3.35/8.77 & 95.00/87.91 \\
   CTC Greedy Search\cite{gravesConnectionistTemporalClassification2006} &   & 3.00/7.59 & 0.33/0.99 & 0.37/0.93 & - & 3.77/9.52 & 94.40/86.93 \\
  CTC Prefix Search\cite{gravesConnectionistTemporalClassification2006}  &   & 3.05/7.58 & 0.32/0.98 & 0.38/0.95 & - & 3.75/9.50 & 94.42/87.05 \\
      \midrule
 Attention\cite{chan2016listen}  &  \multirow{4}{*}{GPT-4o} & 2.43/6.00 & 0.73/1.19 & 0.47/1.12 & -0.18/-0.47 & 3.64/\textbf{8.32} & 95.67/90.40 \\
   Attention Rescore\cite{zhang2020unified}  &   & 2.47/6.39 & 0.25/0.73 & 0.46/1.26 & -0.16/-0.39 & \textbf{3.19}/8.38 & \textbf{96.05}/\textbf{90.10} \\
 CTC Greedy Search\cite{gravesConnectionistTemporalClassification2006}  &    & 2.63/6.36 & 0.31/0.88 & 0.49/1.20 & \textbf{-0.34}/\textbf{-1.07} & 3.43/8.45 & 95.57/90.08 \\
CTC Prefix Search\cite{gravesConnectionistTemporalClassification2006}  &     & 2.75/6.30 & 0.32/0.89 & 0.51/1.22 & -0.17/\textbf{-1.09} & 3.58/8.41 & 95.55/89.82 \\
      \bottomrule
    \end{tabular}
\end{table*}
\begin{table}[!ht]
    \centering
    \caption{CER and Noun Recall of GPT-4o on AISHELL-2 Streaming ASR Error Correction. The ASR model is a conformer-based AED model trained on AISHELL-2 and decoded with the Attention Rescore Decoding Method using a chunk size of 16.}
    \label{aishell2}
    \renewcommand{\arraystretch}{1.25} 
    \newcolumntype{S}{>{\small}c} 
   \begin{resizebox}{1.0\columnwidth}{!}{
    \begin{tabular}{S|SSS|SSS}
      \toprule 
       \multirow{2}{*}{\textbf{\makecell{LLM \\ Correction}}} &
      \multicolumn{3}{|c|}{\textbf{Error Metrics}} & 
       \multicolumn{3}{|c}{\textbf{\makecell{Performance  Metrics}}}
   \\
      \cmidrule(lr){2-4} 
      \cmidrule(lr){5-7} 
     & \textbf{sub} & \textbf{del} & \textbf{insert} & \textbf{Change} & \textbf{CER} & \textbf{Noun Recall}\\
      \midrule
   --&   5.26 & 0.17 & 0.14 & - & 5.57 & 77.64 \\
      \midrule
  GPT-4o  &  4.47 & 0.29 & 0.19 & \textbf{-0.62 (11\%)} & \textbf{4.95} & \textbf{79.66} \\
      \bottomrule 
    \end{tabular}
}\end{resizebox}
\end{table}

We select three datasets as evaluation benchmarks, including the Chinese datasets AISHELL-1~\cite{buAISHELL1OpensourceMandarin2017} and AISHELL-2~\cite{duAISHELL2TransformingMandarin2018}, and the English dataset LibriSpeech~\cite{panayotovLibrispeechASRCorpus2015}, aiming to evaluate performance on both Chinese and English data. The LibriSpeech test set is partitioned into two subsets: \emph{test-clean}, containing high-quality audio, and \emph{test-other}, comprising noisier audio samples.

\subsection{The Model Configuration}
\label{Model Setting}
In our experiment, we adopt a conformer-based attention-encoder-decoder~(AED) ASR model trained with the WeNet toolkit~\cite{yaoWeNetProductionOriented2021}, following the U2++\cite{wu2106u2++} method, across three datasets. For AISHELL-1 and LibriSpeech, a non-streaming model architecture is employed, and four decoding strategies are evaluated: Attention, Attention Rescoring, CTC Greedy Search, and CTC Prefix Search. 

To further assess the performance of streaming models, experiments are conducted on AISHELL-2 using the Attention Rescoring decoding method with a chunk size of 16. The Conformer encoder comprises 12 blocks with an attention dimension of 256, 4 attention heads, and 2048 linear units, incorporating relative positional encoding and Swish activation. The decoder is implemented as a bidirectional Transformer with three forward and three backward layers. The network follows a hybrid CTC/attention architecture, with the CTC loss weight set to 0.3.

For LLMs, we select GPT-4o~\cite{openaiGPT4TechnicalReport2024} and DeepSeek-V2~\cite{deepseek-aiDeepseekv2StrongEconomical2024}, both demonstrating strong multilingual capabilities in Chinese and English. The decoding temperature is set to 0.2, and the top-$p$ parameter is set to 0.8.


\begin{table*}[!ht]
    \centering
    \caption{Framework Analysis Results on AISHELL-1 with DeepSeek v2 and Attention Decoding Method; The ASR model is a conformer-based AED model trained on AISHELL-1}
    \label{table:framework-analysis}
    \renewcommand{\arraystretch}{1.3} 
    \newcolumntype{S}{>{\small}c} 
    \newcolumntype{C}{>{\centering\arraybackslash}X} 
    \begin{tabularx}{\textwidth}{c|c|c|c|C|C|C|c}
      \toprule
      \multirow{2}{*}{\textbf{LLM Correction}} & 
      \multicolumn{3}{c|}{\textbf{Error Metrics}} & 
     \multicolumn{3}{c|}{\textbf{Performance Metrics}}  &
      \multirow{2}{*}{\textbf{Cost Tokens (in/out)}} \\
      \cmidrule(lr){2-4}
      \cmidrule(lr){5-7}
      & \textbf{sub }& \textbf{del} & \textbf{insert}  &\textbf{CER} &\textbf{Change} & \textbf{Noun Recall}& \\
      \midrule
      - & 4.65 & 0.29 & 0.12 & 5.06 & - & 79.32 & - \\
      \midrule
      + Base & 9.76 & 4.56 & 38.76 & 53.10 & +48.04 (949\%) & 55.12 & 62k/89k \\
      + Pre-Detection & 4.71 & 1.62 & 2.10 & 8.19 & +3.13 (61.8\%) & \textbf{81.71} & 72k/87k \\
      + Chain of Thought Sub-Tasks & 5.41 & 0.95 & 0.68 & 7.05 & +1.99 (39.3\%)& 72.30 & 239k/122k \\
      + Iterative Correction & 4.02 & 0.63 & 0.27 & 4.89 & -0.17 (3.3\%) & 77.71 & 251k/261k \\
      + Answer Verification & 4.00 & 0.49 & 0.20 & \textbf{4.69} & \textbf{-0.39 (7.7\%)} & \textbf{81.05} & 251k/260k \\
      \bottomrule
    \end{tabularx}
\end{table*}

\subsection{Experiment Results}
\label{Experiment Results}

Table~\ref{aishell1} summarizes the results on the AISHELL-1 dataset, where GPT-4o achieved an average CER reduction of 0.89 percentage points (17.4\% relative reduction), with a maximum reduction of 1.11 percentage points (21\%). As shown in Table~\ref{aishell2}, GPT-4o achieved the better performance on the AISHELL-2 dataset, with a CER reduction of 0.62 percentage points.

As shown in Table~\ref{librispeech}, on the LibriSpeech test-clean subset, GPT-4o achieved the lowest WER of 3.19\%, corresponding to an average reduction of 0.21 percentage points (5.72\% relative reduction) and a maximum reduction of 0.34 percentage points (9\%). On the test-other subset, a WER of 8.32\% was achieved, with a maximum relative reduction of 1.09 percentage points (11.4\%). 

Experiments with DeepSeek-V2 on LibriSpeech yielded suboptimal performance, as DeepSeek-V2 is primarily designed for Chinese and demonstrates limited capability on English datasets.

\subsection{Result Analysis}
\label{Result Analysis}

The results indicate that LLMs effectively correct substitution errors in speech recognition across both Chinese and English datasets. However, a slight increase in deletion and insertion errors is observed, primarily due to hallucinations. 

To analyze noun recall within substitution errors, noun filtering was performed using the method proposed in~\cite{che-etal-2021-n} for Chinese (AISHELL-1/2) and~\cite{bird2009natural} for English (LibriSpeech). The analysis reveals an improvement in noun recall following correction.

%% file: src/analysis.tex
\section{Ablation Study}
\label{Analisis}

The ablation study analyzes the contribution of different components in \ref{Framework Analisis}, presents a case study of the correction process in \ref{Process Situation Analysis} on AISHELL-1 using DeepSeek-V2, and briefly analyzes token consumption in \ref{Cost Tokens}.

\subsection{Framework Analysis}
\label{Framework Analisis}

To evaluate the effectiveness of each framework component, a component-wise ablation study was conducted on AISHELL-1 using DeepSeek-V2. Results are summarized in Table~\ref{table:framework-analysis}. Consistent with prior findings~\cite{minExploringIntegrationLarge2024}, using LLMs with simple prompting for error correction often leads to increased error rates, as reflected by the baseline results.

To mitigate redundancy and instruction execution errors in LLM outputs, an error pre-detection module was introduced, reducing the CER to 8.19\%. Incorporating CoT subtask decomposition further reduced the CER to 7.05\%, as it primarily constrains the LLM output space and prevents unwarranted word insertions or deletions. However, correct inference cannot always be achieved in a single pass. 

By integrating iterative reasoning into the framework, the CER was reduced to 4.89\%. Finally, implementing answer verification further reduced the CER to 4.69\%, effectively mitigating hallucination-induced errors in LLM corrections.

\subsection{Process Situation Example}
\label{Process Situation Analysis}

\begin{figure}[!ht]
\centering
\includegraphics[width=0.50\textwidth]{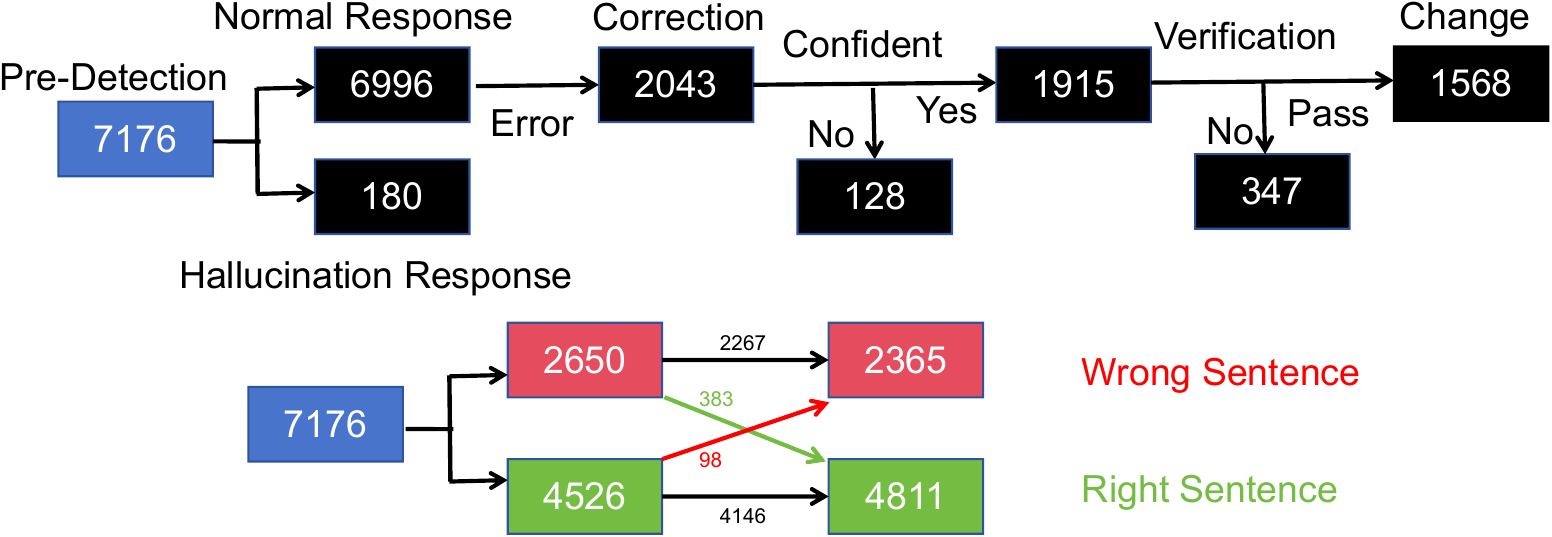}
\caption{Analysis of Results by Sentence on AISHELL-1 Using DeepSeek-V2 and Attention Decoding Method; The ASR model is a conformer-based AED model trained on AISHELL-1}
\label{Anylisis by sentence change}
\end{figure}

As shown in Figure~\ref{Anylisis by sentence change}, the LLM identified 2,043 problematic sentences. Through iterative refinement, confidence was retained for 1,915 sentences, while 128 were discarded and 347 were rejected by the answer verification module. Ultimately, 1,568 sentences were successfully modified. Among these, 383 erroneous sentences were corrected, whereas 98 originally correct sentences were mistakenly altered to erroneous forms.

For sentences containing multiple errors, the overall error count was reduced, although precise quantification remains challenging.

\subsection{Token Consumption Analysis}
\label{Cost Tokens}

To reduce input token consumption during experiments, multiple sentences were grouped together for inference. Empirical observations suggest that this strategy only slightly degrades performance. The specific token consumption is summarized in Table~\ref{table:framework-analysis}. 

Since LLM outputs are inherently variable, the number of output tokens also fluctuates across runs; thus, the reported values serve only as approximate references.

%% file: src/conclusion.tex
\section{Conclusion}
\label{Discusion}

This paper focused on the application of general LLMs to the specialized domain of speech recognition error correction. We proposed the Reliable LLM Correction Framework, which use Pre-detection and Answer Verification to mitigate the LLM hallucination problem and ensure the reliability of LLM answers. Also, We  decompose the error correction task into several subtasks and incorporates  iterative inference and confidence assessment. Our framework enhanced the reliability of LLMs in correcting ASR errors and we achieved better ASR result on representative datasets.